\documentclass[letterpaper, 10 pt, conference]{ieeeconf}  %

\IEEEoverridecommandlockouts                              %

\usepackage{graphicx}
\usepackage{caption}
\usepackage{caption}
\usepackage{subcaption}
\usepackage{amsmath,amssymb,enumerate}

\usepackage{amsthm}
\usepackage{setspace}
\usepackage{amsmath,amssymb,enumerate}
\usepackage{booktabs}
\usepackage[usenames,dvipsnames,svgnames,table]{xcolor}
\usepackage{mathtools}
\usepackage{algorithm, algorithmicx, algpseudocode}

\usepackage{blindtext}
\usepackage{gensymb}
\usepackage{xparse}
\usepackage{lipsum}
\usepackage{mathrsfs}
\usepackage[mathscr]{euscript}
\usepackage{times}
\usepackage[noadjust]{cite} %
\usepackage{multicol}
\usepackage{multirow}
\definecolor{darkgreen}{rgb}{0,0.6,0}
\usepackage[bookmarks=true,colorlinks=true,pdfpagemode=UseNone,citecolor=black,linkcolor=black,urlcolor=BrickRed,pagebackref]{hyperref}
\usepackage{amsfonts}
\usepackage{cleveref}
\usepackage[utf8]{inputenc}
\usepackage[T1]{fontenc}
\usepackage{textcomp}
\usepackage{arydshln}
\usepackage{tabu}
\usepackage{cuted}
\usepackage{balance}
\newtheorem{problem}{Problem}

\definecolor{note}{rgb}{0.1,0.1,1}
\definecolor{rephase}{rgb}{0.15,0.7,0.15}
\definecolor{bag}{rgb}{0.6,0.6,0.2}

\makeatletter
\renewcommand*\env@matrix[1][c]{\hskip -\arraycolsep
  \let\@ifnextchar\new@ifnextchar
  \array{*\c@MaxMatrixCols #1}}
\makeatother

\DeclareDocumentCommand{\vector}{ O{} }{\mathrm{vec}(#1)}

\makeatletter
\newcommand{\mathleft}{\@fleqntrue\@mathmargin0pt}
\newcommand{\mathcenter}{\@fleqnfalse}
\makeatother

\title{\LARGE \bf Providers-Clients-Robots: Framework for spatial-semantic planning for shared understanding in human-robot interaction}

\author{Tribhi Kathuria, Yifan Xu*, Theodor Chakhachiro*, X. Jessie Yang, and Maani Ghaffari \\
\thanks{Funding for M. Ghaffari was in part provided by NSF Award No. 2118818.}
\thanks{The authors are with the University of Michigan, Ann Arbor, MI 48109, USA. {\tt\small\{tribhi, yfx, teochiro, xijyang, maanigj\}@umich.edu}}%
\thanks{*Equal contribution}
}

\begin{document}

\maketitle
\thispagestyle{empty}
\pagestyle{empty}

\begin{abstract}

This paper develops a novel framework called Providers-Clients-Robots (PCR), applicable to socially assistive robots that support research on shared understanding in human-robot interactions. Providers, Clients, and Robots share an actionable and intuitive representation of the environment to create plans that best satisfy the combined needs of all parties. The plans are formed via interaction between the Client and the Robot based on a previously built multi-modal navigation graph. The explainable environmental representation in the form of a navigation graph is constructed collaboratively between Providers and Robots prior to interaction with Clients. We develop a realization of the proposed framework to create a spatial-semantic representation of an indoor environment autonomously. Moreover, we develop a planner that takes in constraints from Providers and Clients of the establishment and dynamically plans a sequence of visits to each area of interest. Evaluations show that the proposed realization of the PCR framework can successfully make plans while satisfying the specified time budget and sequence constraints and outperforming the greedy baseline.

\end{abstract}

\IEEEpeerreviewmaketitle

\section{Introduction}
\label{sec:intro}

Socially Assistive Robots (SARs) are robotic agents, embodied and non-embodied, that are employed to assist humans through social interactions \cite{SAR}. As robotic technology has sophisticated over the years, we have started seeing the acceptance and need of robots in a social space, meaningfully interacting with people \cite{socila_robots}. For robots interacting with people in a service setting, the balance lies in learning behaviors that are social and engaging and sharing expertise to assist the clients \cite{brandao2021towards}. This presents a challenge in two aspects. One, creating a representation of the environment that the robot can use to communicate with the interacting parties. Two, using this expertise to plan the robot behavior.

\begin{figure}[t]
    \centering
    \includegraphics[width=\columnwidth]{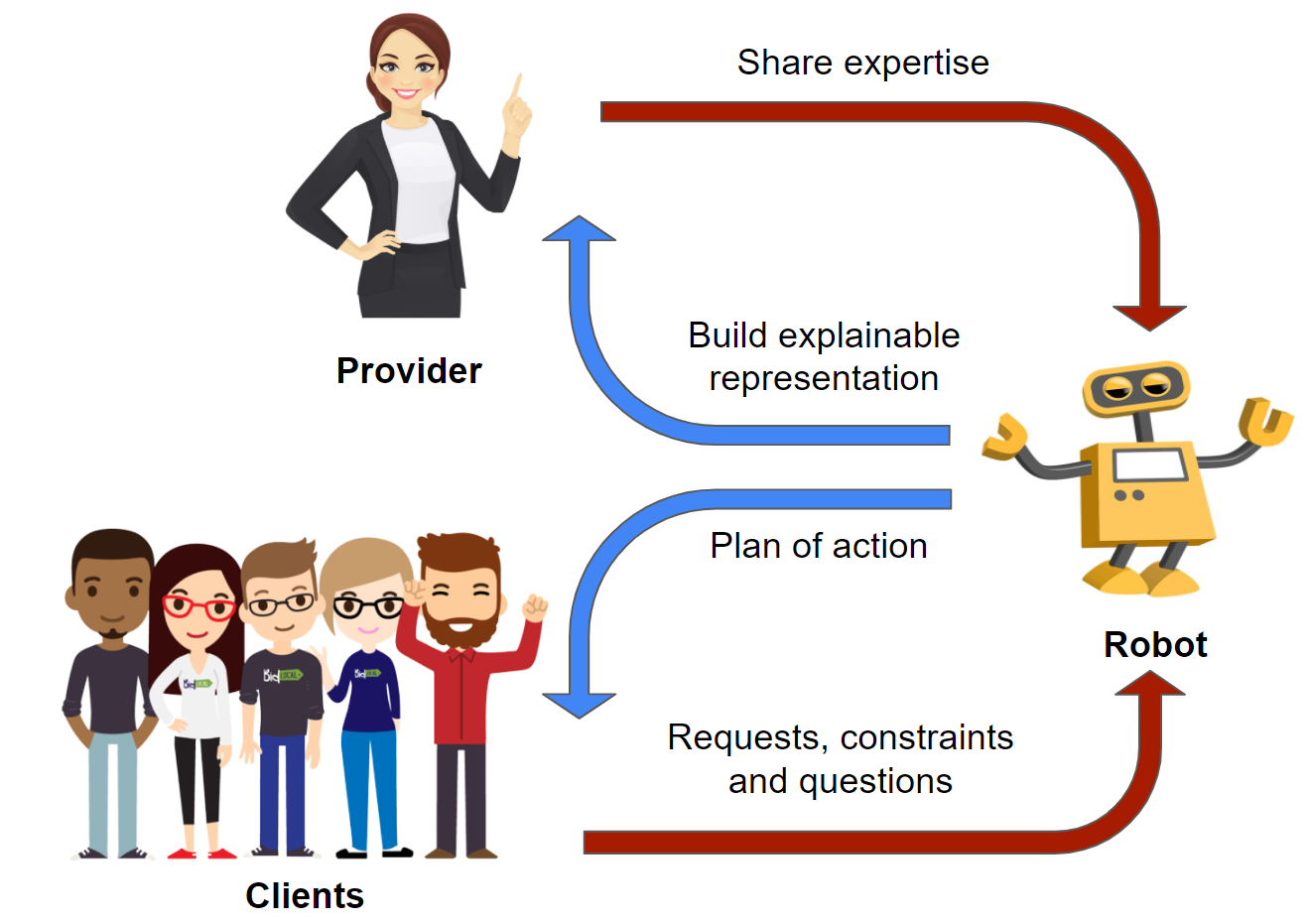}
    \caption{Provider-Client-Robot Relationship.}
    \label{fig:pcr}
\end{figure}

The work of \cite{bartneck2020human} breaks down Human-Robot-Interaction (HRI) in three major parts. These are, spatial interaction, non-verbal interaction and verbal interaction. For this work, we focus on spatial interaction. spatial interaction, refers to the robot and the human sharing and reasoning about the same physical space \cite{HRSI}, \cite{hrsi_qtc}, \cite{mead2012probabilistic}. For a robot operating autonomously in an environment, mapping and navigation are basic components of spatial interaction. Mapping is a way to create a shared spatial understanding. Common map representations explored in robotics are usually occupancy maps and topological maps \cite{thrun2002robotic}. Occupancy maps simply represent an environment as free space and obstacles \cite{thrun1999experiences}. For building human understandable map representations and plans, a widely deployed idea is to use topological maps that represent an environment as a semantically connected graph of the free space \cite{routemodel}, \cite{kuipers1982map}. While these are good works for creating a shared spatial understanding using mapping, for planning these works treat people as obstacles and execute pre-planned actions \cite{thrun1999minerva}, \cite{satake2009approach}. Whereas, works in HRI that focus on planning interactively have found value in assigning roles and setting  expectations of human participants in interaction and planning robot behaviour to meet these expectations \cite{brandao2021towards}, \cite{ong2008implementation}, \cite{explainable_planning}. As a result, there is a gap in research where we have been treating map representations, planning and interactive systems in isolation. 

This work proposes a novel architecture recognizing three interacting parties, Providers (people in-charge at an establishment), Clients (visitors of the establishment), and Robots (PCR) that are involved in a socially assistive setting. As seen in Fig.~\ref{fig:pcr}, we describe the different roles and expectations of interactions between PCR parties. We present an instance of a spatial interaction pipeline within a PCR framework for applications in socially assistive robots detailing a mapping and planning pipeline with the context of a robot in indoor spaces such as an office building, museum, art gallery, or shopping mall. The proposed pipeline successfully uses a shared understanding of the environment for improved and interactive planning. Within the PCR framework, we solve the planning problem as one of understanding and encoding preferences from the Providers and the Clients into planning Robot behavior. 

\begin{figure}[t]
    \centering
    \includegraphics[width = \columnwidth]{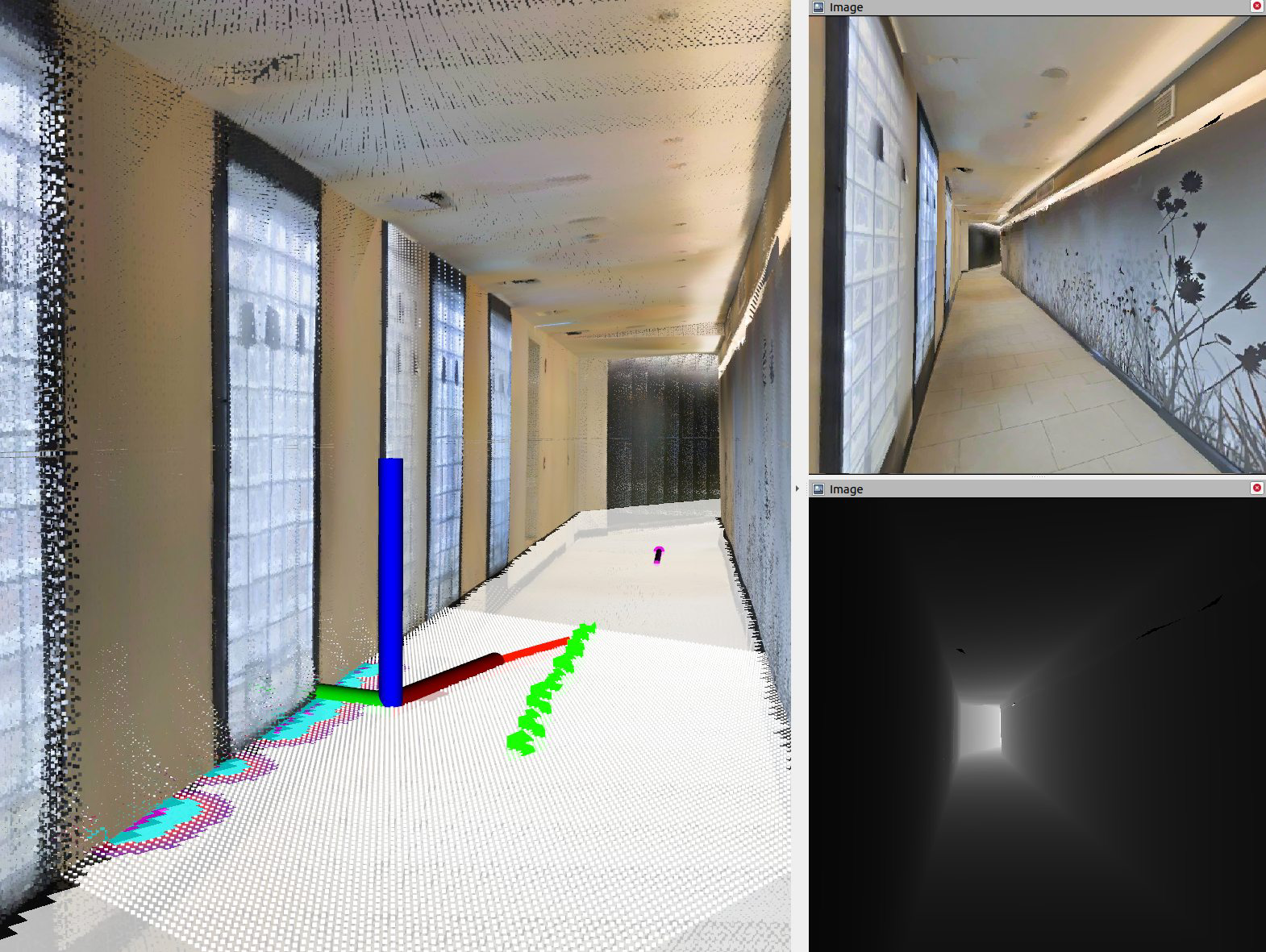}
    \caption{Our robot navigating in a 3D rendered Spa facility in Habitat\_Sim Environment, with the RGB and Depth image on the right panel, and the 3D Map cloud on left. The robot uses a pruned voronoi grid built on top of the topological graph of the 2D map of the environment. The global plan constructed by this voronoi planner is depicted in green. The local planner uses the 3D point-cloud information generated by the RGB-D point-cloud, path can be seen in red. The tour planner plans tours in real-time with this representation of the free space of the environment and specifies the goal to be reached to the planning pipeline.}
    \label{fig:habitat_navigating}
\end{figure}

As shown in Fig.~\ref{fig:habitat_navigating}, we use a photo realistic indoor environment in Habitat-Sim \cite{savva2019habitat} to get results of our implementation in giving tour of a Spa facility, and prepare a pipeline in Robot Operating System (ROS)\cite{quigley2009ros},\cite{robotstack} that can directly be implemented on a real robotic system. 

In particular, this work has the following contributions.

\begin{enumerate}
    \item We develop a novel framework for socially assistive robots within the context of a robot at an establishment and identify Provider and Client roles to set expectations for interaction between them as given in Fig.~\ref{fig:pcr} 
    \item We develop a mapping and planning pipeline within the context of our PCR framework.
    \begin{enumerate}
        \item Mapper: Spatial semantic hierarchical map representation that informs our planning architecture. %
        \item Planner: Linear Programming based tour planner that can encode competing objectives given to the robot from the clients and the providers. %
    \end{enumerate}
    \item The system has been implemented in a ROS \cite{quigley2009ros} pipeline that operates in a simulation environment built from real indoor scenes such as houses and malls. The software is available for download at 
    \url{https://github.com/UMich-CURLY/spatial_interaction}
\end{enumerate}

The remaining of this paper is organized as follows. Related work, discussing a brief history of tour planning and using different maps to plan, is given next.
Our proposed framework is discussed in detail in Section~\ref{sec:methodology}, which describes choosing a map representation in Section~\ref{subsec: mapping} for global planning and, Section~\ref{subsec: planning} that provides our Linear Programming-based approach for tour planning. 

Simulation results of experiments conducted in Habitat-Sim environment are given in Section~\ref{sec:results}. We conclude our results and discuss the limitations and expected future work in Section~\ref{sec:discussion} and Section~\ref{sec:conclusion}. 

\section{Related Work}
\label{sec:relatedwork}

\begin{figure*}[t]
    \centering
    \includegraphics[width = 0.99\textwidth]{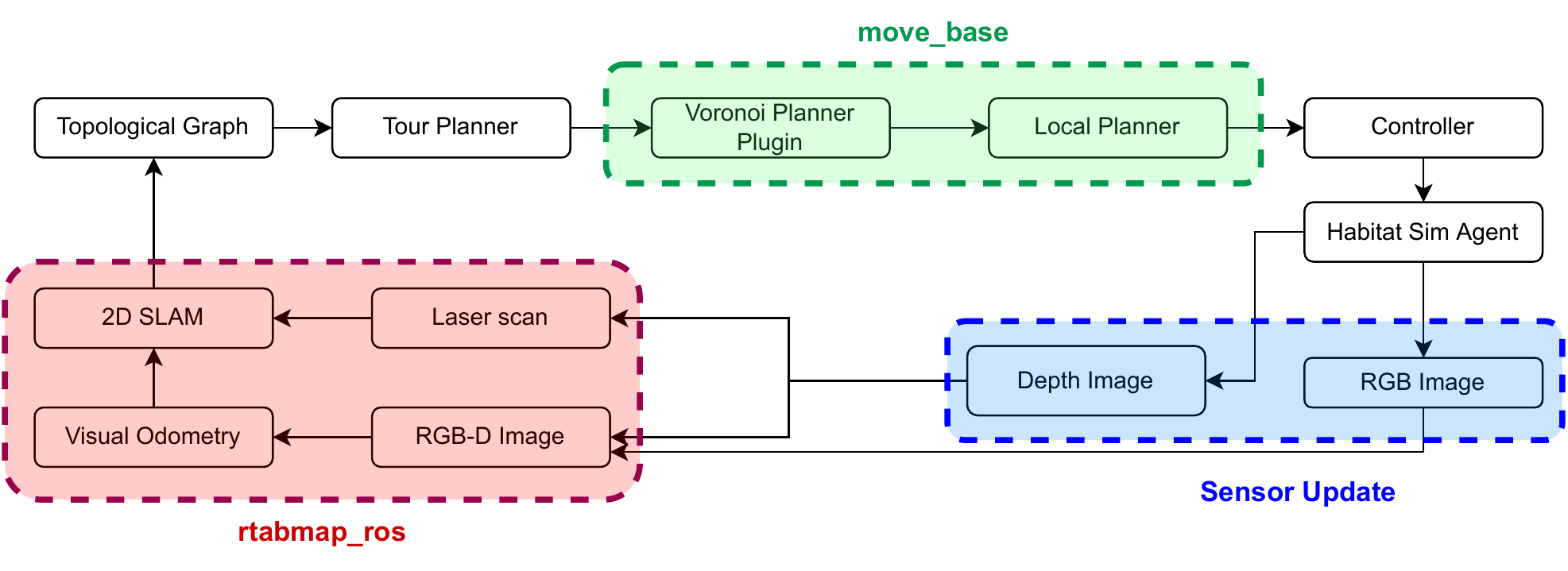}
    \caption{The Mapping + Planning Architecture for an RGB-D agent in the Habitat-sim environment.}
    \label{fig:my_label}
\end{figure*}
For robots cooperating with humans in the scene, it is essential to learn to balance out different needs and requirements. A popular method for categorizing these expectations from Human-Robot Interactions comes from assigned roles to the humans and the robots in the scene. For \cite{ong2008implementation} these roles are based on the amount of human intervention in robot control formulated as Master-Slave, Supervisor-Subordinate, Partner-Partner, Teacher-Learner, and Fully Autonomous. More recently, \cite{brandao2021towards} have been looking towards explainable planning and the different explanation required for the different roles of the humans in the scene (Environment designer, Developer, Lay user, and Mechanical engineer). These approaches either focused on building a shared map representation or explainable planning, but not both together. We identify Provider-Client-Robot roles for socially assistive settings and present a mapping and planning pipeline that uses a shared, and actionable map representation to cohesively plan the robot's navigation behavior. 

For building a shared spatial understanding, studies have looked at classic SLAM problem in context of HRI. A popular example is robots being used to give tours of museums. RHINO and Minerva were some of the early robots giving tours of museums to its visitors. The robot was responsible for executing pre-planned tour given the occupancy map of the environment it was in, and navigate without hitting any obstacles or getting lost~\cite{thrun1999minerva},~\cite{thrun1999experiences}. Following the success of these, researchers started looking at ways to plan for robot behavior based on observed gestures, gaze and speech \cite{bennewitz2005towards}, \cite{faber2009humanoid}.

Most of the research in this field has seen pre-planned tours being executed by robots while they are navigating the environment, treating people, as well as all objects as obstacles. This is a direct result of the chosen map representation, metric maps. While metric maps are good for obstacle avoidance, for most robotic problems, they fail to capture the connectivity between different parts of the mapped spaces.
We propose to have a layer of a topological map that helps the robot find paths between different parts of the building \cite{kuipers2000spatial}. 

\cite{gemignani2016interactive} use topological mapping to demarcate different parts of a house by learning to identify different rooms in it and then evaluate the effectiveness of using topological maps over metric maps by having the robot follow user commands in both cases for comparison, \cite{bastianelli2013knowledge}. As expected, people gave instructions that identified rooms that the topological map could provide a knowledge base for. 
A similar result was seen in an urban search and rescue operation, where a robot was teleoperated by a human for scoping an area for a victim, and was issued commands that required topological maps \cite{casper2002workflow}. Finally, \cite{garcia2015vision} give a survey of several vision based methods that are used today to generate these detailed maps

Mapping ontologies form the knowledge base for a robotic system. Another line of research looks into extracting object-level semantic knowledge from the environment. \cite{Rosinol20icra-Kimera} develop a dense scene graph of their environment, recognising different objects and creating and tracking object meshes. They further learn to segment their environment into different rooms and floors, and establish container-ship of all the sensed object meshes.

We use Spatial Semantic Hierarchy as the representation of our environment, where we choose to augment only areas of interest (AOIs) to our map and treat all other objects as obstacles \cite{johnson2018topological}. Within the context of a PCR relationship, the augmented map is built offline by the Providers and robots, that is then used to assist the clients in the scene. Spatial Semantic map makes a sparse graph, where all points in the corridor amount to a single connected edge for instance \cite{kuipers2000spatial}. As a result this allows flexible, constraint-based, fast tour planning and re-planning. 

Additionally we propose a planning architecture that builds on top of our chosen map representation to dynamically plan and execute tours subject to constraints from both provider and clients. Tour planning when looked at as a variation of the travelling salesman problem (TSP) has been studied extensively in operations research and solved as a Mixed Integer Linear Programming (MILP) problem. \cite{tours} studied the selective TSP to customize tours for visitors trying to explore a new city. The approach is based on building a user profile and using that to define the utility of different points of interest for the TSP. The selective travelling salesman problem which is combination of the simple TSP and the knapsack problem has also been studied in literature as an Orienteering Problem (OP) \cite{LAPORTE1990193}. OPs have been used to solve for planning data collection using aerial vehicles \cite{aerial}, multi-robot planning \cite{shi2020robust}, exploration with multiple robots and survival constraints \cite{OP}. While our previous work looked into a multi-robot tour guiding problem, focusing on matching human teams to robots,  this work develops the single robot tour planning with Provider and Client constraints on top of an actionable map representation \cite{fu2021simultaneous}.

The unique contribution of this paper is an integrated mapping and planning framework that is centered around the identified Provider and Client roles in a socially assistive setting described in Fig.~\ref{fig:my_label}. The map representation encodes information about the different AOIs that are pertinent to the robot planning and execution as determined by the Providers. The MILP based tour planner allows flexibility in encoding Provider and Client preferences and adding constraints easily in real-time. The global planner gets the voronoi paths from the topological graph and the local planner then avoids dynamic obstacle and executes the tour in real time. 

\section{Methodology}
\label{sec:methodology}
Our PCR framework is based on using map representations to inform planning architecture for robots, that is amenable to handling constraints from Providers and their Clients. We note that for making plans that can take into account constraints, we need a shared representation of the environment between all parties, PCR.

As shown in Fig.~\ref{fig:my_label}, we also report on a planning pipeline that makes use of a light-weight, semantic topological map. Our tour planner uses the map representation to encode the connectivity of different areas of interest in our environment to generate optimal plans. The tour is then executed as the robot navigates the environment, getting the voronoi path from our plugin. This trajectory is then sent to the local planner that forward simulates our agent's constant acceleration model, keeping in mind the kino-dynamic constraints, using the Trajectory rollout in the ROS Navigation stack. The mapping and global planning pipeline is described in~\ref{subsec: mapping}. The interactive tour planner built on top is described in detail in Section \ref{subsec: planning}.

\subsection{Mapping}
\label{subsec: mapping}
Our approach to selecting a mapping methodology is motivated by two main ideas. One, for a robot with fixed mobile base and height, a projected 2D map that can ignore all the obstacles above robot height is a good approximation of its obstacle space. Two, when moving in a known environment, it is safe to assume that the structure of the building will not change in the robot's window of operation. We leverage these assumptions to develop a layered mapping architecture that adequately captures the different layers of semantic information, pertinent to our robot's planning. 

\begin{figure}
    \centering
    \includegraphics[width =0.48 \textwidth]{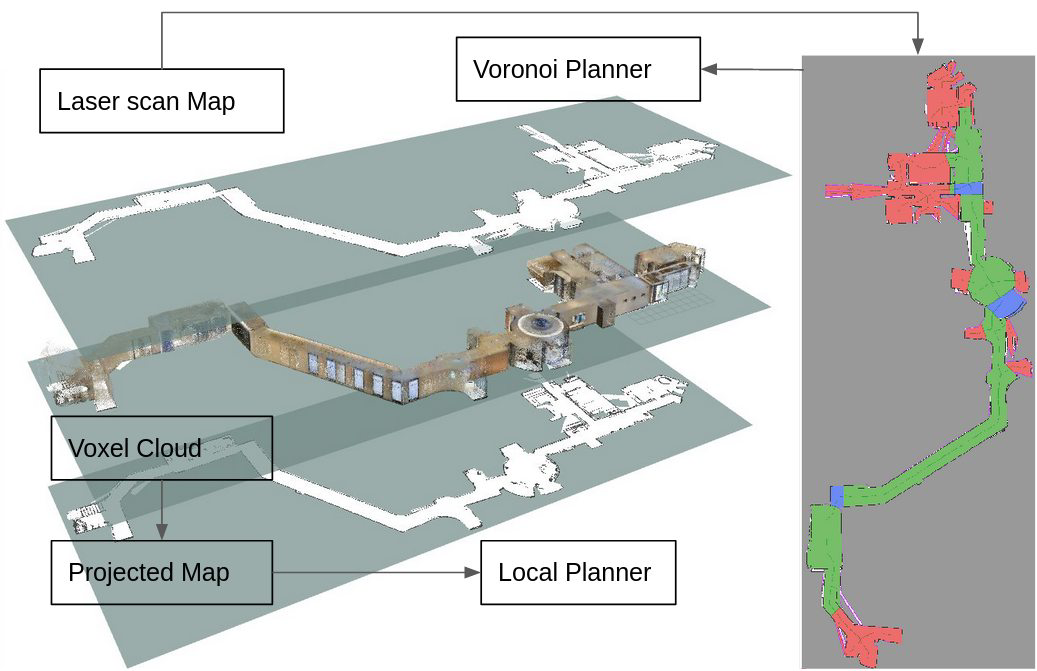}
    \caption{The different layers of our Map. The topological map (right) is built from the occupancy map created of the environment. Green are pathways, Blue are decision points, Red are destinations. There are parts that were not mapped in the environment as the mesh was not fully reconstructed, and were ignored in laser scan and projected map. The voxel cloud helps the local planner avoid any dynamic or floating base obstacles not included in the sparse laser scan map.
    }
    \label{fig:map_layers}
\end{figure}

For experiments in Habitat-Sim simulation environment, our agent only has on-board an RGB-D camera. We capture the connectivity of the regions in our space by building a spatial semantic hierarchy on the occupancy grid, obtained by simulating sparse laser from our depth image as depicted in Fig.~\ref{fig:map_layers}.
The environment is then represented as a collection of path segments and gateways. Furthermore, the areas formed in the map are then labelled as either pathways, places or decision points. Pathways are connecting areas, such as hallways, corridors and the like. Decision points are at the intersection of pathways, where a decision can be made to follow a path that enters different areas. Finally, a place is the end-points of the graph, usually a room, a bathroom and so on. A voronoi skeleton describes the connectivity between these different areas in the environment.

The semantic hierarchy is built on features detected from our 2D laser field of view, described in detail in \cite{kuipers2000spatial}, \cite{johnson2018topological}. To avoid obstacles that may have either a floating or thin base such as tables, chairs and, shelves we use a projected 2D map, cut off at our robot height. For real-world environments, such as the one captured in Matterport3D dataset, this is essential to make sure we are not planning paths that may have a collision with different parts of the robot's body. We use the rtabmap\_ros package in ROS to get both the projected 2D map with and without 3D dynamic obstacles \cite{labbe2019rtab}.  

Finally, we use our mapping algorithm to create tours and execute them using a global planner that follows the voronoi path between the start and goal pose in the tour. We write our plugin to the standard move\_base package in ROS navigation stack that can query these paths from our skeleton. We compare the time to plan tour using our map representation against a simple move\_base planner running A* on the projected map. The results for the same are discussed in Section~\ref{sec:results}.

\subsection{Planning}
\label{subsec: planning}
This section describes the integration of our map representation and its implementation to plan and execute tours (or other assistive tasks). The planning constitutes two parts. First is the high-level Tour planner that takes in constraints from both the Providers and Clients of the establishment and plans the place and sequence of visits to each object/area of interest. The second is a voronoi planner that helps provide the paths that can connect these different locations, followed by the local planner. 
The tour planner gets the poses where the robot should be to be able to see the objects/areas of interest in the scene and uses the voronoi paths between different paths of the topological graph to get an estimate of the travel time between all the different poses in the map. Finally, the planner returns a sequence of robot poses, that are visited in order. We recognize this as a variation of the traveling salesman problem and solve it using a MILP. 

\begin{table}[t]
    \centering
        \caption{Variable Definition.}
    \footnotesize
    \resizebox{\columnwidth}{!}{\begin{tabular}{cccc}\toprule
     Variable & Definition \\ \midrule
        $x_{ij}$ & Defines if the edge $\{i,j\}$ is visited, $\forall \{i,j\} \in (1,N)$\\ \midrule
        $y_{i}$ & Defines if the $i$-th node is visited  \\ \midrule
        $a_{li}$ & Is true if client $l$ wants to visit node $i$ \\ \midrule
        $L$ & Set of all clients in the scene   \\ \midrule
        $M$ & Set of all nodes \\ \midrule
        $N$ & Total Number of edges \\ \midrule
        $S$ & Set of sequence constraints \\ \midrule
        $T_{i}$ & Time spent visiting the $i$-th node\\  \midrule
        $T_{ij}$& Time to travel between node $i$ \& $j$ \\ \midrule
        $T$ & Total time for execution of tour \\ \midrule
        $T_{lim}$ & Time limit of execution of tour \\ \midrule
        $t_i$ & Time of reaching any node i \\ \midrule
        $T_{L}$ & A constant large amount of time \\ \midrule
        $w_{d}$ & Defines the weight of the dropped demand cost \\ \midrule
        $w_{t}$ & Defines the weight of the Time cost \\ \bottomrule
    \end{tabular}}
    \label{tab:tourplanner}
\end{table}

Table \ref{tab:tourplanner} defines the different variables that are used in the definition of our optimization problem. The cost function can then be written as given in \eqref{eq: cost}: 
\begin{equation}
   \mathcal{C}(y_{i}, T) = {w_d}  \sum_{l \in L} \sum_{i \in N} a_{li}(1-y_i) + w_{t}T
    \label{eq: cost}
\end{equation}
There are two competing objectives that we are trying to optimize. The first is the dropped demand cost which penalizes plans where the robot decides to drop a node requested by a client. The other is the time objective, where we try to minimize the total time of the tour, thus resulting in the shortest total tour between the chosen nodes of visit. Generally, the weight on the demand objective is significantly higher to make sure that the agent satisfies the maximum amount of demands while staying within the constraint of a time limit.
Next, we define the bounds of our variables in \eqref{eq: graph bounds} and \eqref{eq:visit time bound}. We should note here that our decision variables, $x_{ij}$ and $y_{i}$ are binary variables that determine if the edge or node, respectively, are visited or not. Finally, \eqref{eq:visit time bound} makes sure that the time spent at any node $i$ is greater than 0. 
\begin{align}
    x_{ij}, y_{i}  \in \{0,1\} &\quad \forall \{i,j\} \in N \;, \; \forall l \in L  \label{eq: graph bounds}\\
    T_{i}  >0  &\quad \forall \{i,j\} \in N \;, \; \forall l \in L \label{eq:visit time bound}
\end{align}
Now that we have the definitions for our variables ($x_{ij}$ \& $y_{i}$) in place, we set our network flow constraints in \eqref{eq:network_1}, \eqref{eq:network_2} Here we ensure, that for every place of visit $m$, the net flow of number of edges should be null. Finally, \eqref{eq:network_2} ensures that the total number of edges coming into any node $m$ is less than 1, implying that each node is visited at most once. 
\begin{align}
    &\sum_{i \in N} x_{im} =  \sum_{j \in N} x_{mj}, \; \forall m \in M \label{eq:network_1}\\
    & \sum_{i \in N} x_{im} \leq 1 \label{eq:network_2}
\end{align}
Finally, we allow our algorithm to accept hard constraints that the provider might require for operation. An example of these can be time constraints on the total length of the tour, and sequence constraints, where we can specify dependencies of one node on the other. For example, in an office space, a client might request to be shown a particular office, but to get to the office the client must first wait in the waiting room (encoded in the node time) and then proceed to the office. 
The total time to complete the tour can be calculated as given in~\eqref{eq: time total}.
\begin{equation}
    T = \sum_{i \in N} \sum_{j \in N} T_{ij}x_{ij} + \sum_{i \in N} T_{i}y_{i}   \label{eq: time total}
\end{equation}
Equations \eqref{eq: time in} and \eqref{eq: time out} constraint the time of reaching any node $j$ to be larger than the time of arrival at $i$, adding on the time $T_{ij}$ to go from $i$ to $j$ and the time $T_i$ spent at $i$. Finally, \eqref{eq: time limit} limits the total time of the tour $T$ to be less than the specified threshold $T_{lim}$.
\begin{align}
    t_{i} - t_{j} + T_{ij}+T_{i} &\leq T_{L}(1-x_{ij}) \label{eq: time in}\\
    t_{i}-t_{j}+T_{ij}+T_{i} &\geq -T_{L}(1-x_{ij}) \label{eq: time out}\\
    T &\leq T_{lim} \label{eq: time limit}
\end{align}
Lastly, the sequence constraint for any pair $\{i,j\} \in S$ are defined in \eqref{eq: seq1} and \eqref{eq: seq2}. 
\begin{align}
    y_{i} &\geq y_{j} \label{eq: seq1} \\
    t_{i} &\leq t_{j} + T_{L}(1-y_{i}) \label{eq: seq2}
\end{align}
Finally, we encode our tour planning setup into a MILP, and use a Google OR-tools solver \cite{or-tools}. We compare the performance of our tour planner against two greedy baselines, as reported in Section~\ref{sec:results}. The resultant optimization problem that minimizes the cost defined in \eqref{eq: cost} is summarized as follows:
\begin{problem}[Tour Planning]
\label{prob:tour}
\begin{align}
 &\min_{\textnormal{$y_i, T$ }} \ \
 {w_d}  \sum_{l \in L} \sum_{i \in N} a_{li}(1-y_i) + w_{t}T \label{eq: minimize} \\
  &   \textnormal{subject to} \ \ \eqref{eq: graph bounds}-\eqref{eq: seq2}. \nonumber 
 \end{align}
\end{problem}

Given the nodes, also referred to as the AOIs, to be visited in the scene, we can formulate the tour planning problem as a MILP as described in Problem ~\ref{prob:tour}. The solution of our tour planner is a sequence of nodes/AOIs that should be visited. As depicted in Fig.~\ref{fig:habitat_navigating}, the sequence of AOIs, is then sent to our voronoi planner plugin. The global planner then extracts a set of start and a goal from the sequence AOIs, returns a voronoi path between the current start and goal that the local planner can follow. The local planner is a kino-dynamic planner that forward simulates the differential drive dynamics of our robot to safely navigate between the start and the goal given 3D obstacle point clouds extracted from our RGB-D cameras. 

For our implementation, we have chosen a photo-realistic physics based simulator called Habitat-Sim \cite{savva2019habitat}, and built a ROS wrapper around it that can plan and execute tours \cite{chen2021ros}. To easily resolve package dependencies for new install, both ROS and Habitat package were built in Conda environments \cite{robotstack}.

\section{Results}
\label{sec:results}

For testing our proposed framework we deploy an agent in the physics based Habitat-Sim environment. Habitat-Sim is the front-runner for simulated embodied navigation. It sources data from real-world environments that can be collected using a Matterport-3D camera. We deploy an agent in this environment to be able to give tours of a spa facility as shown in Fig.~\ref{fig:habitat_navigating}. We select a total of 20 poses that the robot must visit to see the objects of interest in the scene such as paintings, sauna chairs, chandelier in the lobby and so on.

Our tour planner has a hard constraint on finishing the tours within a pre-determined time. As a result for optimal planning, we need to compute the distances between every two nodes (AOIs) in addition to the distance between all nodes and the current robot position. Our mapping methodology has an inbuilt graph-like structure and can return the distances between these nodes efficiently.
Finally, adding the current pose of the robot and AOIs to the scene, we only need to look at the distance between that point and the closest point of connection to our topological graph and that returns the path distance between the start and goal node. This is necessary when we are planning tours over time as with the number of AOIs in the scene the number of path computations increases exponentially. We compare the time of planning tours using our voronoi planner on top of the topological graph against planning over an occupancy grid using a grid-based planner such as A*. As is shown in Table~\ref{tab:voronoiplanner} our chosen map representation improves planning time by 81\%-85\%.

\begin{table}[h]
    \centering
        \caption{Runtime of voronoi vs. A* Trajectory Planner. From the runtime table, the time running voronoi path planner is 81\%-85\% less than normal A* move\_base path planner}
    \footnotesize
    \resizebox{\columnwidth}{!}{\begin{tabular}{ccccc}\toprule
Trajectory Planner	& \multicolumn{4}{c}{Runtime $(s)$}\\ \midrule
                    &	20 Nodes & 30 Nodes & 40 Nodes	& 50 Nodes \\ \midrule
move\_base, A*      &	8.174	 &17.102	 & 30.262	& 47.686 \\ \midrule
Voronoi Plugin       &	\textbf{1.516}	 & \textbf{2.742}	     & \textbf{4.897}	& \textbf{7.394} \\ \midrule
\% improvement      &   81.453	 & 83.967	 & 83.817	 & 84.449 \\ \bottomrule
    \end{tabular}}
    \label{tab:voronoiplanner}
\end{table}
\begin{figure}[t]
    \centering
    \begin{subfigure}{.4\textwidth}
    \includegraphics[width = \columnwidth]{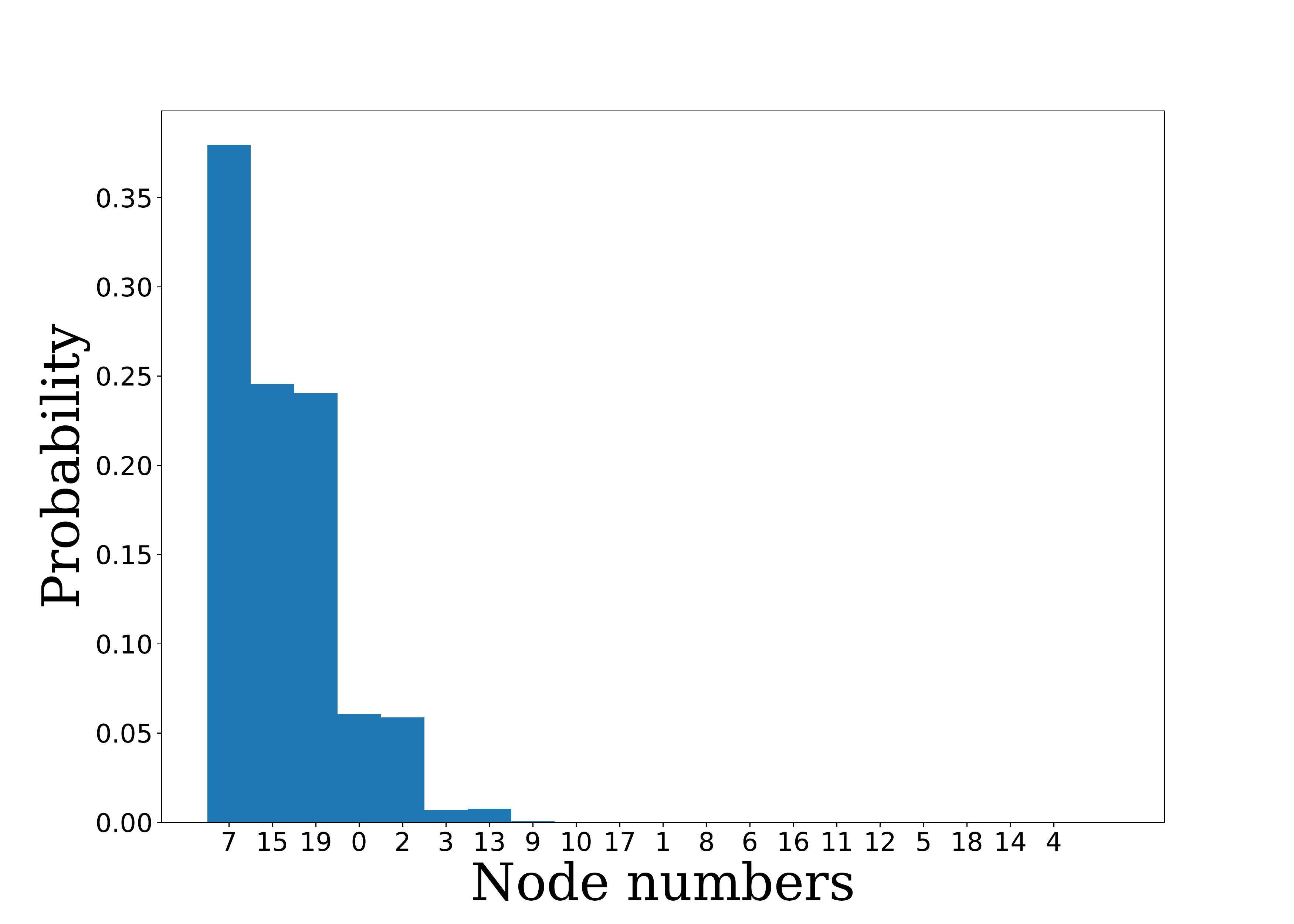}
    \caption{Standard deviation is 1}
    \label{fig:client_choice_dev_1}
    \end{subfigure}
    \begin{subfigure}{.4\textwidth}
    \includegraphics[width = \columnwidth]{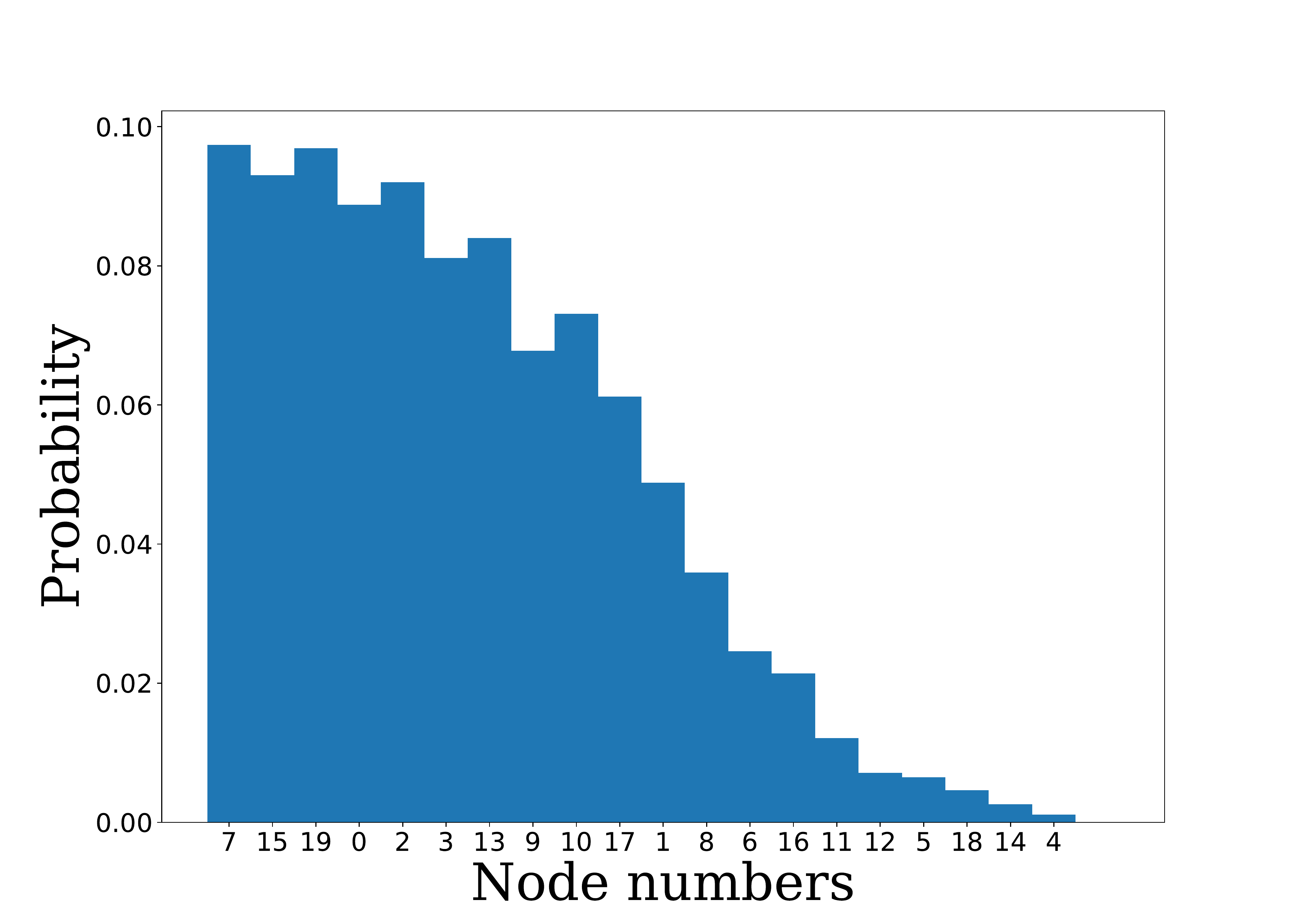}
    \caption{Standard deviation is 5}
    \label{fig:client_choice_dev_5}
    \end{subfigure}
    \caption{Probability of any node [0-19] being selected by the client for different values of deviation. The choices are sampled with replacement, where each client can make at most 5 unique demands. As a result when the deviation is low, there is a strong bias in the selection, and we are sure to see more popular demands repeated.}
\end{figure}

Additionally, our tour planner takes into account different client requests. The client demand is then sampled from the gaussian distribution given in Figures~\ref{fig:client_choice_dev_1} and~\ref{fig:client_choice_dev_5} for different deviation. The AOIs are arranged in the order of popularity, simulating the client choice being more biased to the popular attractions when the deviation is low and less biased when the deviation is high. We use this to design different planning scenarios having a wide range of total demand and compare our performance against two greedy baselines. 

As described by the cost function in equation (\ref{eq: cost}), our planner optimizes on two competing objectives, coming in from the clients and the providers. The planner then tries to maximize the number of client demands that can be fulfilled within the time limit and the sequence constraint allotted by the provider. 

\begin{figure}[h!]
    \centering
    \includegraphics[width = \columnwidth]{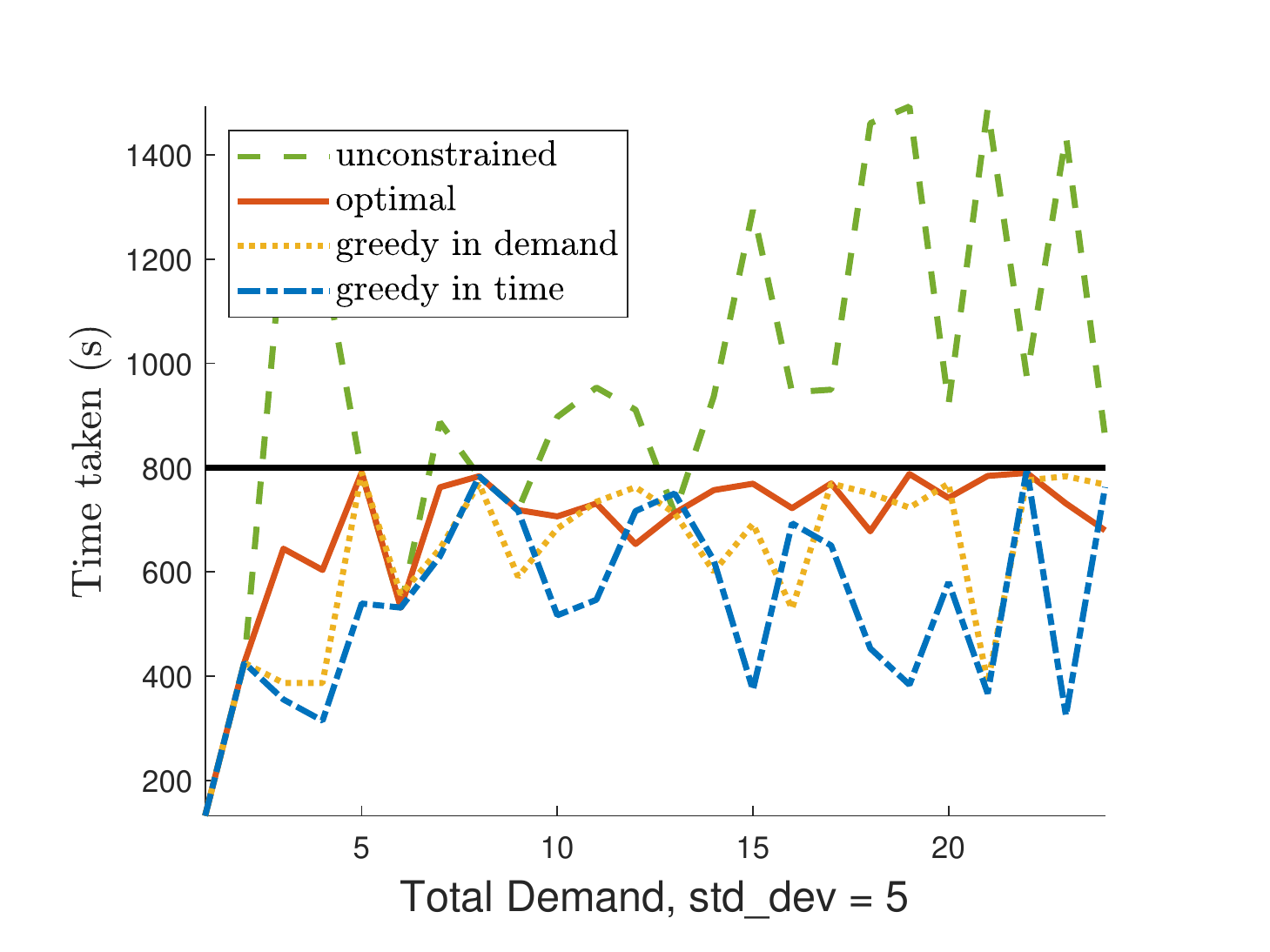}
    \caption{The time taken to complete the tour. The greedy and optimal agents are constrained and have to drop some demands to always stay within the time limit, given in black. The unconstrained agent, covers all the demands, but can go over the required time limit to meet all the demand in shortest possible amount of time. In comparison to greedy baselines, the optimal agent takes the most of amount of time (visits as many places as possible) while staying within the time budget.}
    \label{fig:time_analysis}
\end{figure}

We compare our performance with two greedy baselines. The first one being greedy in time and the other one being greedy in demand. Both planners are set to complete tours within the prescribed time limit, which in this case is set to 800 seconds. The greedy in time agent chooses the nearest node from the current node to the list of nodes that are requested by the client and continues to visit them until there is still budgeted time. Whereas, the greedy in demand agent sorts the list of client demands in ascending order and tries to visit them in that order (visiting the most popular nodes first), dropping the nodes that cannot be visited due to the hard constraints on time. Both planners account for the time to get back to the starting position and do so within the time constraint as can be seen in Fig.~\ref{fig:time_analysis}.

Fig.~\ref{fig:time_analysis} shows the time taken to execute the tour for the optimal and the greedy planners against a naive unconstrained planner that never drops any demand and simply finds the shortest route between the requested AOIs. The time taken is plotted against total demand, which is the non-unique sum of the total number of AOIs requested by [1-10] clients each of whom can choose [1-5] AOIs, and is varied from [1-30]. 

\begin{figure}[ht!]
    \centering
    \begin{subfigure}{\columnwidth}
    \includegraphics[width = \columnwidth]{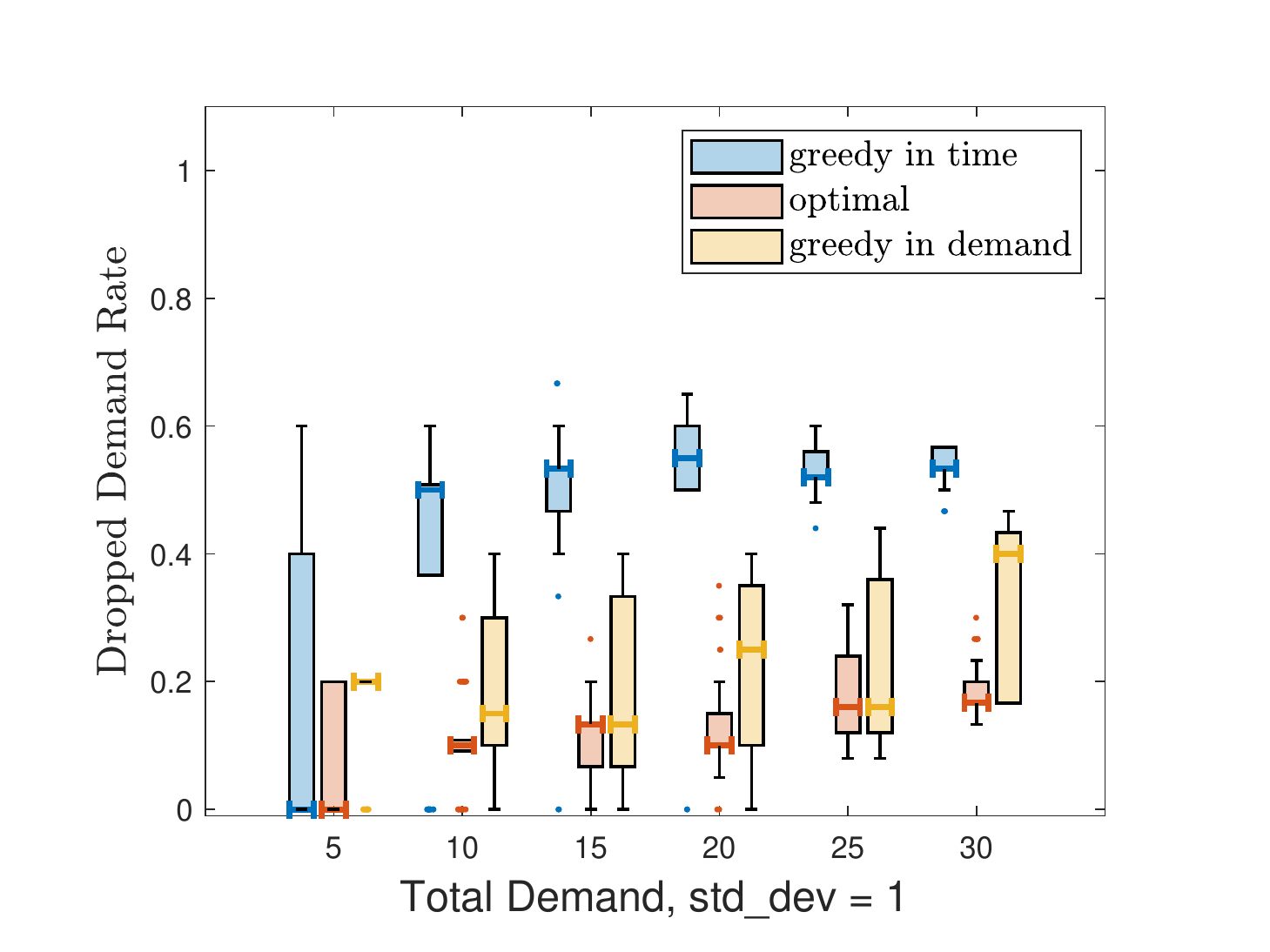}
    \caption{Deviation is low (1), so there were a lot of repeated choices, the greedy in demand agent does better than the greedy in time. The performance of greedy agents is comparable for low demand loads with the optimal but gets worse at higher demand load.}
    \label{fig:demand_vs_client}
    \end{subfigure}
    \begin{subfigure}{\columnwidth}
    \includegraphics[width = \columnwidth]{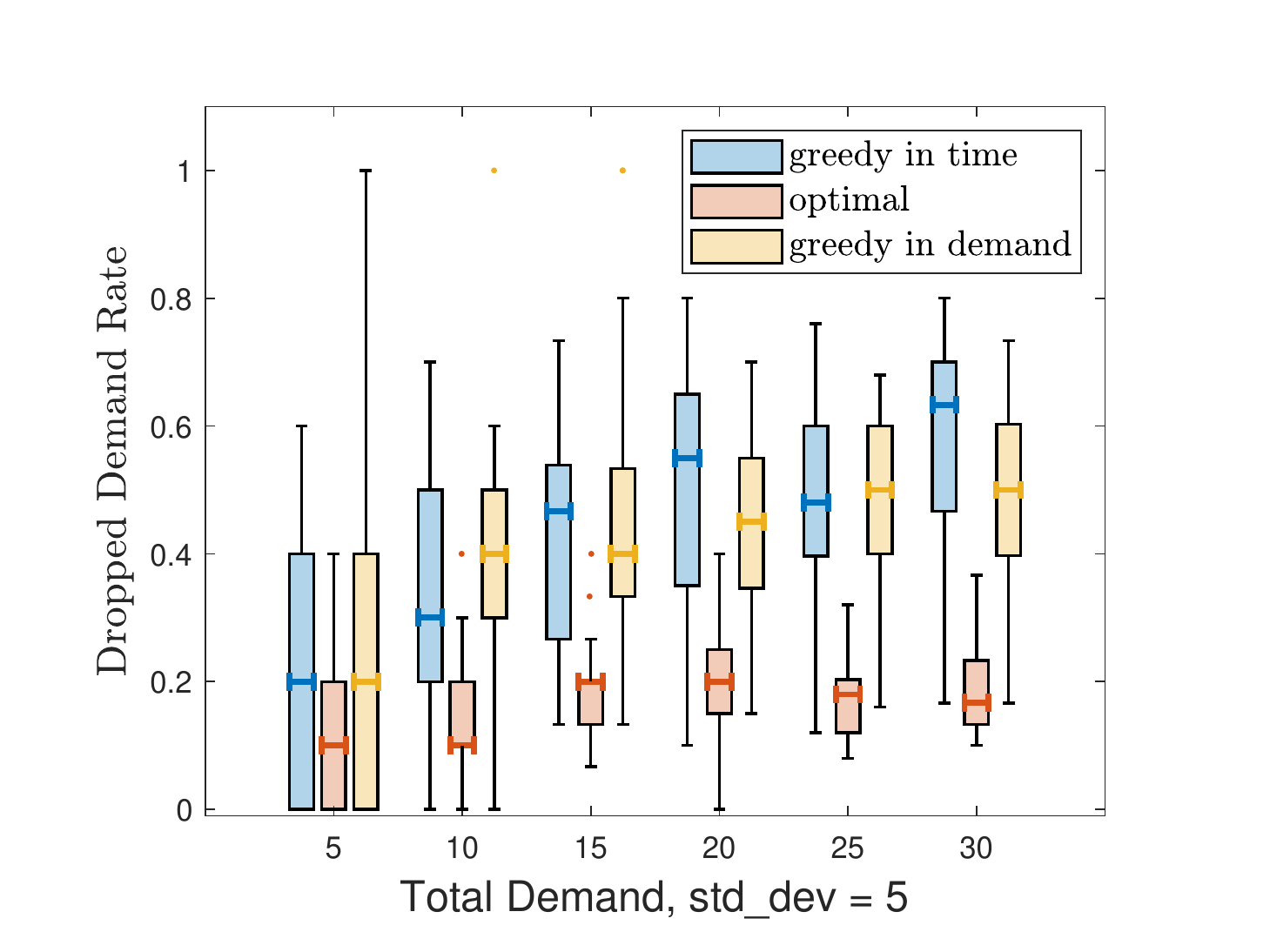}
    \caption{Deviation is high (5), so there was no clear bias in choices, the greedy in demand agent is no longer close to optimal even in low demands. 
    }
    \label{fig:demand_vs_client_high}
    \end{subfigure}
    \caption{Distribution of Demand drop rate for 50 runs with different total demand load each. The ``.'' depicts the outliers, and colored line is the median. For higher deviation the drop rate is higher, but the optimal planner outperforms the greedy baselines, dropping fewer client demands. }
\end{figure}

The demand drop rate is plotted against the total demand in Figures ~\ref{fig:demand_vs_client} and ~\ref{fig:demand_vs_client_high} respectively. 
Where the demand drop rate is described as
\begin{equation*}
    \textrm{demand drop rate}=\frac{\textrm{demands not satisfied}}{\textrm{total demand}}
\end{equation*} 
The total demand is varied in bins of 5 from [5-30] and the demand drop rate is collected for a total of 50 runs each. 
We show a comparison of the two cases when the human choices are more spread out (standard deviation of choice is higher) and less spread out for comparison in Figures~\ref{fig:demand_vs_client},~\ref{fig:demand_vs_client_high}

As can be noted when the demand is low, the optimal planner converges to the greedy planners, as they are able to fulfil the demand greedily within the constraints of time. However, when the number of humans or the variation in their choice increases, our tour planner optimizes by maximizing the client demand while staying within the specified constraints.
A more nuanced analysis can be made in comparison of Figures~\ref{fig:demand_vs_client},~\ref{fig:demand_vs_client_high}. When the deviation in client choices is low in Figure~\ref{fig:demand_vs_client}, i.e. there is a bias in client choice and the demand is high but mostly repetitive (everyone wants to see the most popular attractions), the optimal planner is similar to the greedy in demand planner on extremes of demand. This is expected since when the demand is either too low or too high, the demand objective outweighs and thus the greedy in demand agent is similar to optimal. While in Fig.~\ref{fig:demand_vs_client_high}, the greedy in demand agent performs worse due the lack of a clear bias in client demands and the optimal agent is similar in performance to the greedy in time planner in lower demands.

\section{Discussion}
\label{sec:discussion}
The proposed PCR framework is set up to design robotics systems that operate in socially assistive settings. We discussed an instance of the same, focusing on the spatial interaction around the PCR roles at a facility. The proposed pipeline is built around creating shared spatial semantic understanding and collaborative decision making. 
We looked at a simplified setup of expectations that can be seen in Fig.~\ref{fig:pcr} and found that we could successfully develop an effective spatially interactive system, for mapping and planning. 

\begin{figure}
    \centering
    \includegraphics[width = 0.9\columnwidth]{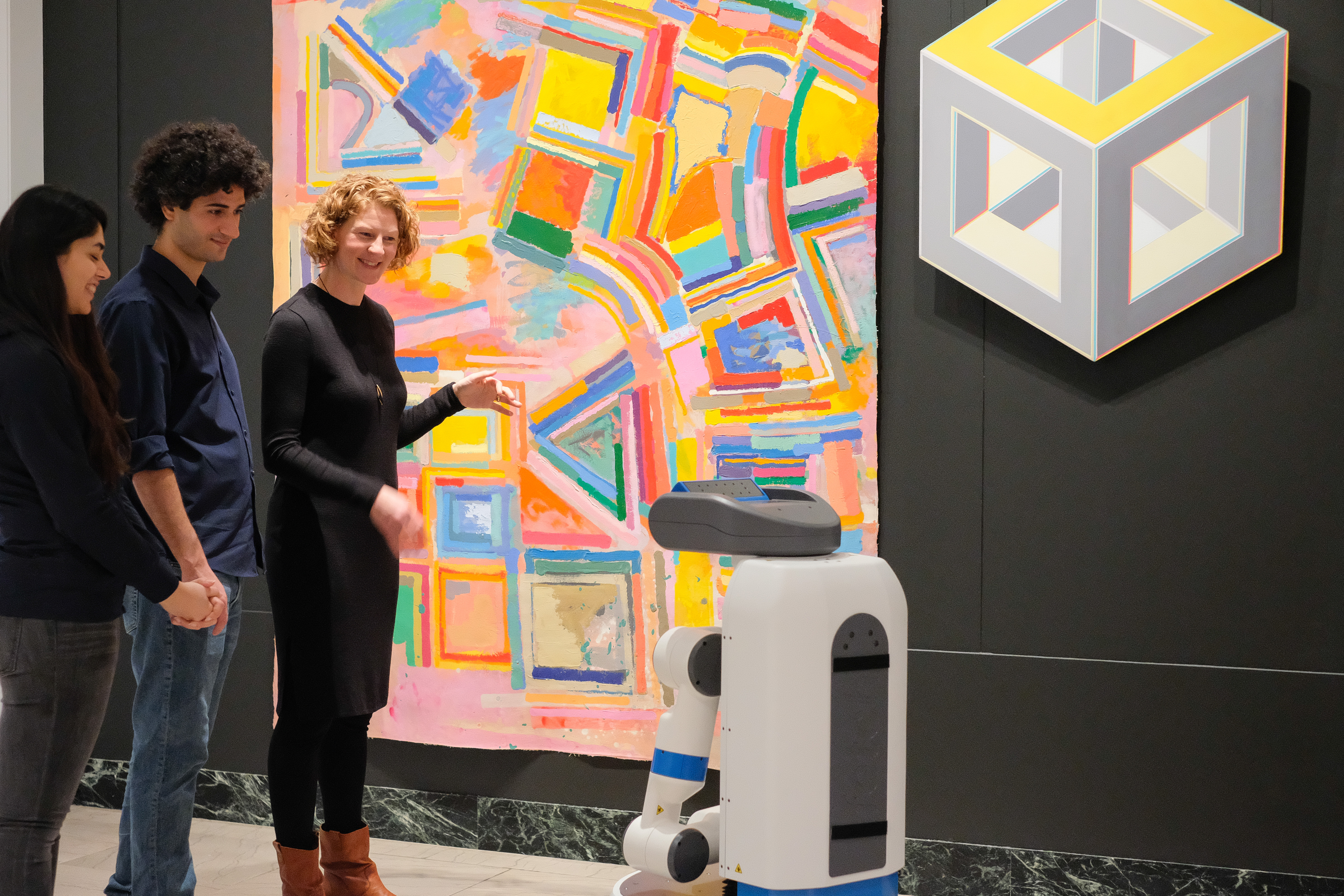}
    \caption{Robot at University of Michigan Museum of Art (UMMA).}
    \label{fig:umma}
\end{figure}

Our insight is that the PCR approach is the appropriate assignment of roles for humans interacting with SARs. 
We want to deploy our robotic system in a real-world environment, and have a long-standing partnership with University of Michigan Museum of Art, where the robot will be used to interact with visitors in the future as seen in Fig.~\ref{fig:umma}. Bridging the gap between the discussed setup and a real-world deployment will be the focus of our future research. We are interested in designing systems of communication that infer inputs from the clients based on interaction. We're also interested in evaluating trust between the interacting parties, in a real uncontrolled setup with a long-term deployment of an autonomous agent \cite{haspiel2018explanations}. 
Finally, we are interested in learning social behavior online, multimodally, and see it's dependence on the semantic understanding between the robots and other dynamic parties in the scene \cite{zhang2021hierarchical}.

\section{Conclusion}
\label{sec:conclusion}

This work proposes a novel Provider, Client and Robot framework for deployment in any general assistive setting. We then go on to detail a spatial interaction pipeline as an instance of the proposed framework. We recognize a need for formulation of navigation around accepting constraints and preferences from both the providers and client, and balance the objective. We share results where our tour planner is compared against greedy agents that don't balance client and provider objectives. We also emphasize the need for an actionable and intuitive mapping representation for a known environment. We directly use our map representation to efficiently inform our planning pipeline to generate better tours. 

In conclusion, we visualize an effective spatial interaction pipeline that is built around a novel PCR framework that can generalize across various service domains. This work is a first step in the direction of a system that sets up stage for interesting future research for a long-term autonomous agent that can interact with the providers and clients at an establishment meaningfully.

\section*{Acknowledgment}
{
The authors would like to thank Prof. Benjamin Kuipers for sharing his insights on this project. He proposed the idea of the Providers-Clients-Robots framework during a meeting. We would also like to thank our collaborators at University of Michigan Museum of Art, Grace VanderVliet and John Turner, for their constant support through our progress. 
}

{\small 
\balance
\bibliographystyle{IEEEtran}
\bibliography{bib/strings-abrv,bib/ieee-abrv,bib/references}
}

\end{document}